\documentclass[conference]{IEEEtran}
\IEEEoverridecommandlockouts
\usepackage{cite}
\usepackage{amsmath,amssymb,amsfonts}
\usepackage{algorithmic}
\usepackage{graphicx}
\usepackage{textcomp}
\usepackage{xcolor}
\usepackage[ruled, linesnumbered]{algorithm2e}
\def\BibTeX{{\rm B\kern-.05em{\sc i\kern-.025em b}\kern-.08em
    T\kern-.1667em\lower.7ex\hbox{E}\kern-.125emX}}
\begin{document}

\title{Structural Pruning in Deep Neural Networks:\\A Small-World Approach}

\author{
\IEEEauthorblockN{Gokul Krishnan,
Xiaocong Du,
Yu Cao}
\IEEEauthorblockA{School of Electrical, Computer and Energy Engineering\\
Arizona State University, Tempe, Arizona 85287\\
Email: \{gkrish19, xiaocong, yu.cao\}@asu.edu}
}

\maketitle

\begin{abstract}
Deep Neural Networks (DNNs) are usually over-parameterized, causing excessive memory and interconnection cost on the hardware platform. Existing pruning approaches remove secondary parameters at the end of training to reduce the model size; but without exploiting the intrinsic network property, they still require the full interconnection to prepare the network. Inspired by the observation that brain networks follow the Small-World model, we propose a novel structural pruning scheme, which includes (1) hierarchically trimming the network into a Small-World model before training, (2) training the network for a given dataset, and (3) optimizing the network for accuracy. The new scheme effectively reduces both the model size and the interconnection needed before training, achieving a locally clustered and globally sparse model. We demonstrate our approach on LeNet-5 for MNIST and VGG-16 for CIFAR-10, decreasing the number of parameters to 2.3\% and 9.02\% of the baseline model, respectively.
\end{abstract}


\section{Introduction}
Recent developments in Deep Neural Networks (DNNs) have made them an integral part of modern day data processing which enable applications such as image recognition \cite{Krizhevsky:2012:ICD:2999134.2999257}, object detection \cite{ren2015faster}, speech recognition \cite{graves2013speech} and other applications. To achieve better accuracy, DNNs are increasingly deeper and wider, resulting in higher cost of memory, interconnect and computation. Such a trend poses a significant challenge to deploy DNNs on a hardware platform. 

To reduce the model size, network pruning, either threshold based \cite{hansong, HaoLi, Zhuang} or loss function based \cite{Baoyuan}, removes secondary weights from a pre-trained model, which is usually over-parameterized. While these methods effectively compress the model, the pruning step does not incorporate the structural property of the network and thus, they cannot control the exact location to be trimmed. Because of that, they still need a redundant network to support all possible connections, when deployed on hardware \cite{kadetotad2016efficient, mohanty2017random, Du}. 

\begin{figure}[t]
\begin{center}
\includegraphics[width=0.95\columnwidth]{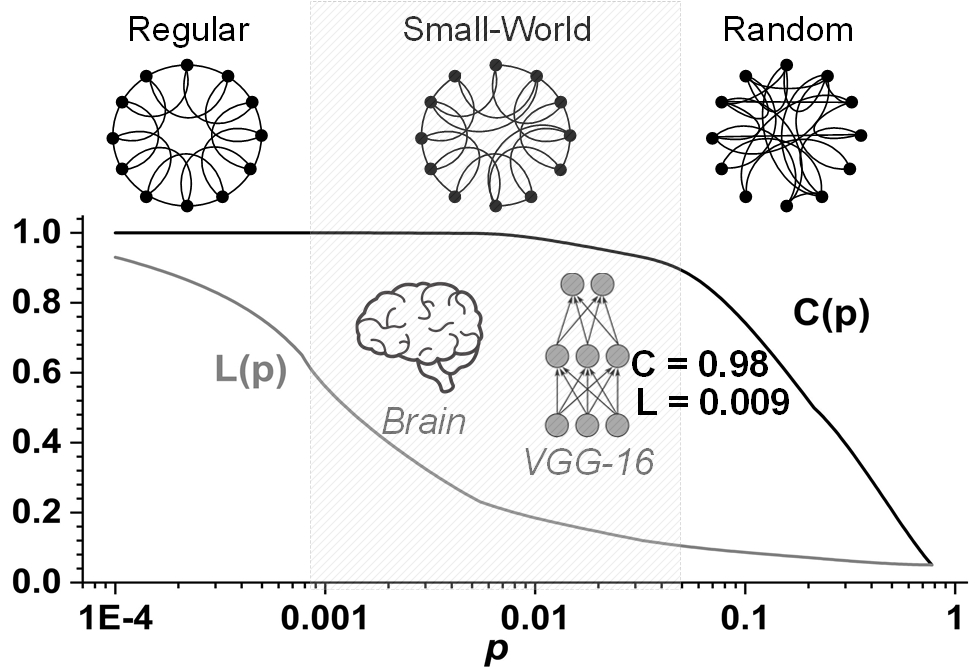}
\end{center}
\vspace{-0.3cm}
\caption{The classical LC curve to illustrate the structural property \cite{Watts1998}. The Small-World region exhibits high Clustering Coefficient C and low Characteristic Path Length L.}
\label{Final_figures2}
\vspace{-0.48cm}
\end{figure}
In this work we aim to address both over-parameterization and the structural redundancy in network reduction. Our inspiration comes from the structural property of the brain model, a complex network both in spatial and time scales \cite{SW_brain}. It is believed that the brain network follows a semi-random structure called the Small-World Network \cite{Watts1998}. The Small-World model can be characterized by two properties, namely, Clustering Coefficient (C) and Characteristic Path Length (L). The Clustering Coefficient gives the cliquishness of a network, i.e., how clustered a network is, and the Characteristic Path Length gives information on the average shortest path length between two nodes in the network. The amount of rewiring that the network is subjected to is based on the random probability $p$ and follows the binomial distribution. Figure.~\ref{Final_figures2} shows L and C under various randomness in connection, ($p$). In the region of a Small-World network, the unique property is that C still stays at a high value close to that of a full network, but L significantly decreases. In DNN, such a property implies high accuracy (i.e., high clustering) with sparse connectivity (i.e., low L). 

Leveraging such a structural property, we propose to train from a sparse Small-World neural network model and enforce the structural constraint throughout the training, instead of training an over-parameterized network from scratch. We demonstrate our approach in image classification tasks, where we evaluate the algorithm for image datasets MNIST \cite{Lecun} and CIFAR-10 \cite{Krizhevsky2009LearningML} with networks LeNet-5 \cite{Lecun} and VGG-16 \cite{vgg} respectively. At the same accuracy, we demonstrate a significant reduction in the number of parameters and computations. For example, on our  implementation of LeNet-5 for MNIST, the Small-World Model reduces the parameters to 2.3\%, lower than \cite{hansong}; on VGG-16 \cite{vgg} for CIFAR-10, the model achieves a reduction in parameters to 9.02\%. To summarize, the key contributions of our work are the following: 

(1) A structural pruning method based on the Small-World model before training, which prevents the model from moving to a local minimum due to over-parameterization and saves the interconnection cost on hardware; 

(2) The combination of both structural properties of a neural network and the importance of parameters to enhance the effectiveness of model reduction; 

(3) A scalable, clustered DNN model which provides the ground for efficient hardware implementation with lower memory cost and sparse interconnection.
\section{Related Work}
DNNs comprise of deep and wide convolutional and fully connected layers that introduce high memory and computation cost on hardware platforms. Reduction of model size to address the hardware cost has been an area of active research for quite some time. We limit the discussion of this section to pruning of networks and Small-World networks.

\textbf{Pruning:} \cite{hansong} performs pruning by removing unimportant connections from a trained model based on weight importance resulting in a network with large number of zeros. Another work \cite{Suraj} introduces gate variables to impose a constraint on the structure to compress it further. \cite{HaoLi} proposes a method to prune channels based on the absolute value of the weights as the metric and retrains the model to regain accuracy. Similar pruning approaches include \cite{Zhuang,Yihui}. Loss function based reduction methods have been explored in \cite{Wei}.

\textbf{Small-World:} Prior work has been done in looking at networks as a Small-World model \cite{Simard, Xiaohu}. But the existing works focus only on feed forward networks and not DNNs. Their approaches focus on rewiring the network for efficient information transfer rather than model compression.

In contrast our work focuses on achieving the goal of a sparse network by using structural properties of a Small-World network to create a sparse structure for DNNs before training and enforcing the structural constraint throughout the process to achieve the final compact and clustered model. The work focuses on achieving high clustering; a property of a regular network; and low shortest path length; a property of a random network. The pruning mechanism is guided and prevents over-parameterization of the model. Additionally, it gives insight into the interconnect reduction achieved layer-wise due to knowledge about the sparsity introduced before training, thus helping reduce hardware cost.
\section{Our Approach}
In this section, we describe the Small-World neural network model proposed to generate the final sparse model for training. We first look into the background on Small-World networks and introduce the  terms we use for this work, followed by the implementation of such a model for a neural network structure.
\begin{figure}[!t]
\begin{center}
\includegraphics[width=1\columnwidth]{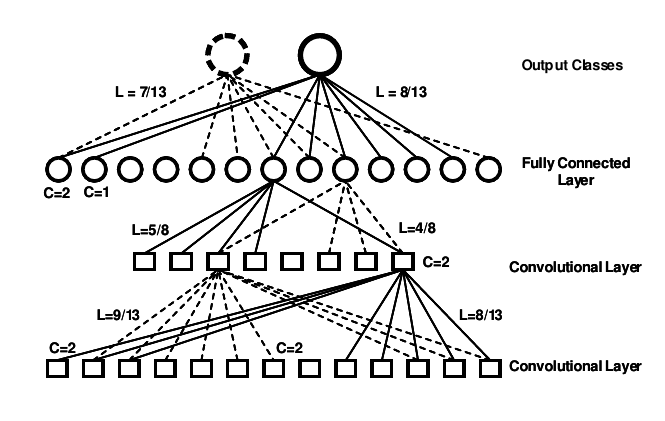}
\end{center}
\vspace{-0.5cm}
\caption{Our methodology to define L and C for a DNN. The Small-World approach produces a locally clustered, globally sparse structure.}
\label{lc_calc}

\end{figure}
\subsection{Small-World Neural Network}
\textbf{Deep Neural Networks:} DNNs are deep networks with large number of layers in them. They consist of two types of layers, namely, convolutional and fully connected layers. Convolutional layers are used as feature extractors and fully connected layers the classifier. Each layer learns the weight and bias parameters through training. Weight parameters of convolutional layers are 4D matrices, $W_l^{channel}\in\mathbb{R}^{op_l\times ip_l\times a_l\times a_l}$, while that of fully connected layers are 2D matrices, $W_l^{i,j}\in\mathbb{R}^{x_l\times y_l}$.

\textbf{Small-World Networks:} Networks that are neither regular nor random are classified as Small-World networks. They exhibit the following properties; (a) high Clustering Coefficient C and (b) low Characteristic Path Length L. Small-World Networks are formed by introducing randomness into a regular network. The randomness is a measure of rewiring performed in the network based on a rewiring probability $p$. Rewiring is the process of removing regular connections and adding long range shortcuts from one node to another as shown in Figure.~\ref{Final_figures2}. This process increases the efficiency of information transfer within the network. The human brain is a classic example of a Small-World Network \cite{SW_brain}.
\begin{table*}
\begin{center}
\resizebox{0.9\textwidth}{!}{ 
\begin{tabular}[t]{|c|l|l|l|}
\hline
\textbf{Label} & \textbf{Name} & \textbf{Description} \\ \hline
$L$ & Characteristic Path Length
  & Number of neurons/feature maps
  contributing to a class, averaged over all classes. \\\hline
$C$ & Clustering Coefficient
  & Number of classes contributed to by a neuron/feature map averaged over all of them \\\hline
$p$ & Random Probability
  & Random rewiring probability following Binomial distribution\\\hline
$\theta$ & Sparsity Metric
  & Percentage of connections to be removed from the network\\ \hline
\end{tabular}
}
\vspace{0.2cm}
\caption{Terminology and notations used in the Small-World based pruning approach for Deep Neural Networks.}
\label{def}
\end{center}
\vspace{-0.4cm}
\end{table*}

The classical Small-World paper \cite{Watts1998} introduces L and C through the example of friendship networks, where L is the average number of friendships in the shortest chain connecting two people and C reflects the extent to which friends of one are also friends of each other. We take cue from this to define L and C for Small-World Neural Networks as:

\textbf{(1)} Friendship is defined as a weighted connection between two neurons/feature maps in a class and each class as a friendship network.

\textbf{(2)} For fully connected layers Characteristic Path Length L is the number neurons connected to the class (friendships per class), averaged over all the classes. Clustering Coefficient C is the number of classes each neuron is contributing to (friends who are also friends of each other), averaged over all neurons.

\textbf{(3)} For convolutional layers Characteristic Path Length L is the number output features contributing to the class (friendships per class), averaged over all the classes. Clustering Coefficient C is the number of classes each output feature contributes to, averaged over all output features.

Table.~\ref{def} gives a comprehensive summary of the terms we introduce with the definitions for our work. We calculate the defined terms L and C of a neural network based on the above methodology as shown in Figure.~\ref{lc_calc}. We show two output classes which are connected to a cluster of neurons. The L value is calculated as the number of neurons connected to the class, averaged over the total number of possible connections. Hence for the first FC layer L is $7/13$ and $8/13$. The C value is calculated per neuron basis, if a neuron contributes to n classes then $C=n$. Hence for the first neuron in Figure.~\ref{lc_calc}, it contributes to 2 classes, thus $C=2$.

A similar approach is adopted for convolutional layers with a difference that the output activations of feature maps are taken to evaluate the contribution to a output feature map. A threshold is chosen based on the mean $\mu$ and standard deviation $\sigma$ of the sum of the absolute values of the activations. Feature maps with activations outside the window $\mu-\sigma$ to $\mu+\sigma$ are dropped. In this manner we quantify the properties of the neural network to enable the final goal of a Small-World neural network model.
\begin {equation}
L_{class} = number\ of\ connections/total \ possible\ connections
\end{equation}
\begin {equation}
L = \Sigma L_{class^i}/\Sigma number\ of\ classes
\end{equation}
\begin {equation}
C_{neuron} = number\ of\ classes\ contributed\ to/total \ classes
\end{equation}
\begin {equation}
C = \Sigma C_{neuron^i}/\Sigma number\ of\ neurons
\end{equation}
\vspace{-0.4cm}
\begin{figure}[!b]
\begin{center}
\includegraphics[width=1\columnwidth]{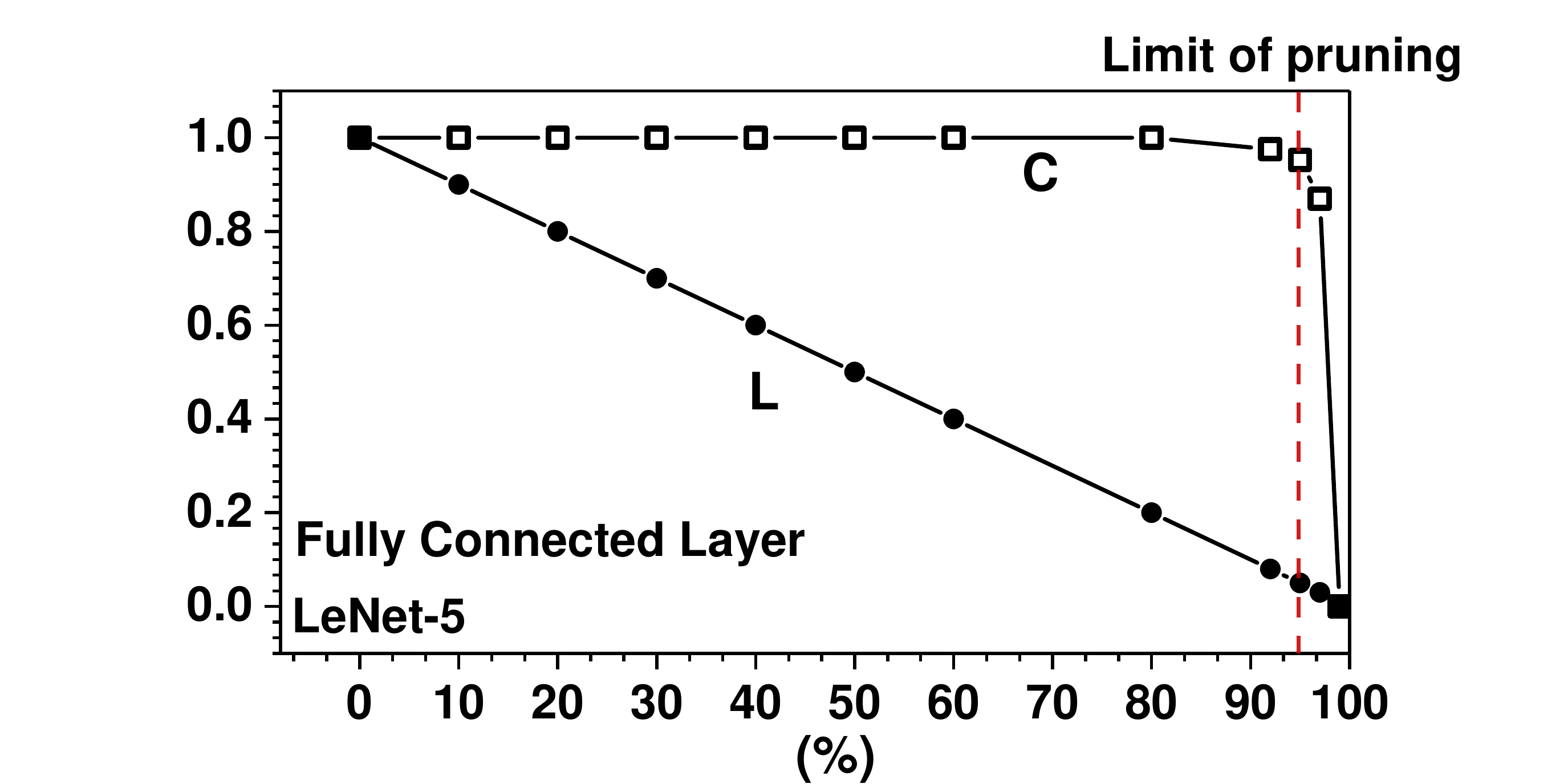}
\end{center}
\vspace{-0.4cm}
\caption{LC Curve for different sparsity using existing pruning methods \cite{hansong}. A regular network can be pruned till the limit when the clustering property disappears.}
\label{lcthresh}
\vspace{-0.2cm}
\end{figure}
\subsection{\textbf{Algorithm}}
The methodology adopted to convert a regular structure into a Small-World model involves two steps, extraction of the network structure parameters and enforcing the Small-World structure onto the network. The algorithm deals with the network layer-wise, modeling each layer as a Small-World model.

\textbf{Extraction Phase:}
This phase deals with the extraction of network structure properties and the addition of randomness into the structure. We define the algebraic model based on \cite{PhysRevE.89.012812} work for regular graph structures. 

Consider a layer L in the regular neural network. Small-World model requires three main parameters for the network, namely, randomness probability, degree of the network and number of nodes in the network. We modify the definition of degree and number of nodes to accommodate the structural nature of a neural network.

Let the number of neurons in a particular layer $l$ be $N_l$ and adjacent layer ${l+1}$ be $M_{l+1}$. The number of nodes from $l$ to $l+1$ is $N_l + M_{l+1}$. The degree of the layer is defined as $M_{l+1}$, which gives the number of interconnections in a regular layer of a neural network. The random probability \textit{p} is user defined and adds randomness to the network. The randomness is calculated through the generation of a binomial random variable based on the random probability $p$ and degree $M_{l+1}$. The number of connections to be rewired is calculated as, $pa$ connections to be removed and $pb$ connections to be added, given by
\begin{equation}
pa = (((N_l + M_{l+1})-M_{l+1}-1)/(N_l + M_{l+1}-1))\times p
\end{equation}
\begin{equation}
pb = ((M_{l+1})/(N_l + M_{l+1}-1))\times p
\end{equation}
Two random variables are generated, $randvar_a$ and $randvar_b$ by drawing samples from the binomial distribution function given by,
\begin {equation}
 f(p) = \dbinom{n}{k} p^k(1-p)^{n-k}
\label{binomial}
\end{equation}
where $p$ is the number of removed/added connections, $n$ is the degree and $k$ the number of successes. On increasing $p$, the binomial distribution approximates to be a normal distribution (Gaussian) at high $n$ thus making the randomness more prominent and resulting in reduction of accuracy as shown in Fig.~\ref{rand}.

The base matrix is split into three sections based on a \textit{divide factor} derived from the degree and the number of nodes as,
\begin{equation}
divide factor = ({M_{l+1}/(N_l + M_{l+1})})\times\textit{p}
\label{div}
\end{equation}
to create two matrices $\Delta a$ and $\Delta b$. The random matrix is generated as,
\begin{equation}
randvar = \Delta a\times randvar_a + \Delta b\times randvar_b
\label{randvar}
\end{equation}

Finally a \textit{randomizer matrix} $\Gamma$ is created using the weight matrix ${W_l}^{i,j}$ and $randvar$ as,\\
\begin{equation}
\Gamma = {W_l}^{i,j} + {randvar} + 2\times {W_l}^{i,j}\times {randvar}
\label{randzer}
\end{equation}

\textbf{Structuring Phase}
\begin{figure}[b]
\begin{center}
\includegraphics[width=1\columnwidth]{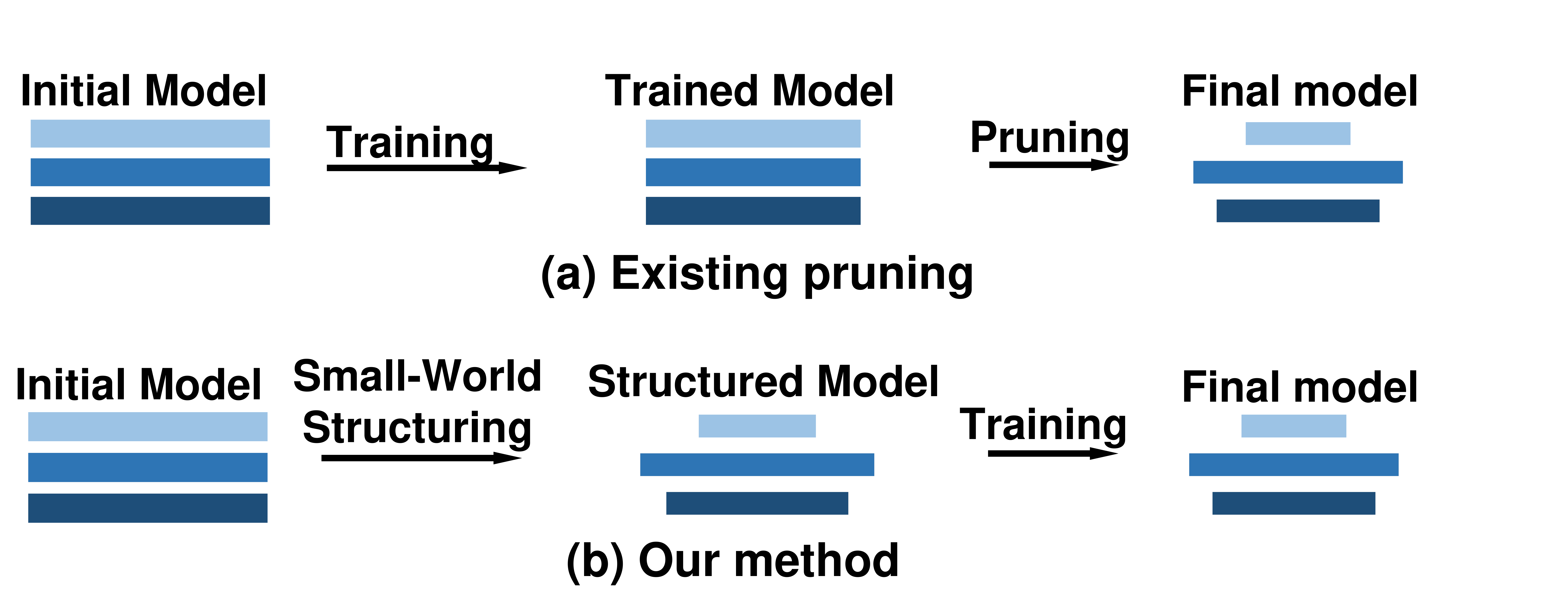}
\end{center}
\vspace{-0.3cm}
\caption{(a) Regular pruning method (b) Proposed Small-World method. The Small-World method uses a sparse network for training and achieves a sparse and accurate model comparable or better than existing methods.}
\label{flow}
\vspace{-0.3cm}
\end{figure}
In this phase the neural network is modeled as a Small-World network by introducing randomness into the weight matrix using the randomizer $\Gamma$. The obtained randomizer matrix $\Gamma$ is then multiplied with the weight matrix ${W_l}^{i,j}$ of layer $l$ to obtain the final Small-World model parameters of layer $l$ of the neural network. 
\begin{equation}
{W_{l_{sw}}} = {W_l}^{i,j}\times \Gamma
\label{final_eqn}
\end{equation}
where ${W_{l_{sw}}}$ is the Small-World weight matrix. The sparse structure is obtained by further pruning this matrix for a particular sparsity level $\theta$. The Small-World model allows the removal of more connections due to structuring of the parameter matrix ${W_l}^{i,j}$ resulting in a more sparse structure to start with.

This is performed for other layers in a similar manner. For convolutional layers the degree is calculated as the size of the output feature map $op_l\times a_l\times a_l$, where $op_l$ is output channel depth and $a_l$ the width and height of the feature map. The number of nodes for the Small-World structure is taken as $op_l\times ip_l\times a_l\times a_l$, where $ip_l$ is the input channel depth.
\begin{algorithm}[t]
\caption{Small-World Structural Pruning} \label{alg:small_world}
\SetAlgoLined
Initialize the Regular Neural Network\\
Extract structure properties of the network layer-wise\\
Calculate number of connections to be removed $pa$ and to be added $pb$\\
Generate the random variable matrix using Equation.~\ref{binomial} \\
Calculate the divide factor as per Equation.~\ref{div}\\
Create the 2 matrices $\Delta a$ and $\Delta b$ from the base matrix based on the divide factor\\
Multiply the number of rewired connections with $\Delta$ to generate the random matrix as in Equation.~\ref{randvar}\\
Compute the randomizer as in Equation.~\ref{randzer}\\
Generate the final Small-World matrix as per Equation.~\ref{final_eqn}\\
Prune the connections based on $\theta$ to generate the sparse Small-World matrix.
\end{algorithm}
\section{Experiments}
This section deals with the experiments performed to validate the Small-World based pruning for Neural Network. Experiments are performed on standard image classification datasets MNIST and CIFAR-10 on DNN architectures LeNet-5 and VGG-16 respectively. The experiments are done using Tensorflow \cite{tensorflow} platform on NVIDIA GeForce TITAN X.
\subsection{Flow of Small-World based Pruning}

This section deals with the process of converting a regular Neural network into a Small-World neural network. Figure.~\ref{flow} gives the flow diagram for traditional pruning and our proposed Small-World methodology. In traditional methods an over-parameterized model is used to start training. Pruning and retraining is done on the trained model to achieve the corresponding final compact sparse model. We perform pruning in such a manner on regular networks and plot the corresponding L and C values for various sparsity metric $\theta$ as shown in Figure.~\ref{lcthresh}. We infer that a sparse model exhibits properties L and C similar to a Small-World network, with high C and low L. We propose a new methodology inspired from this observation where we first model the regular network into a sparse Small-World model and then let it train while imposing the structural constraint on it as shown in Figure.~\ref{flow}(b). This produces an accurate and sparse model with sparsity comparable or better than existing methods. We argue that an over-parametrized network is not a necessity to start  training a neural network so as to achieve an optimum model.
\begin{figure}[!t]
\begin{center}
\includegraphics[width=0.9\columnwidth]{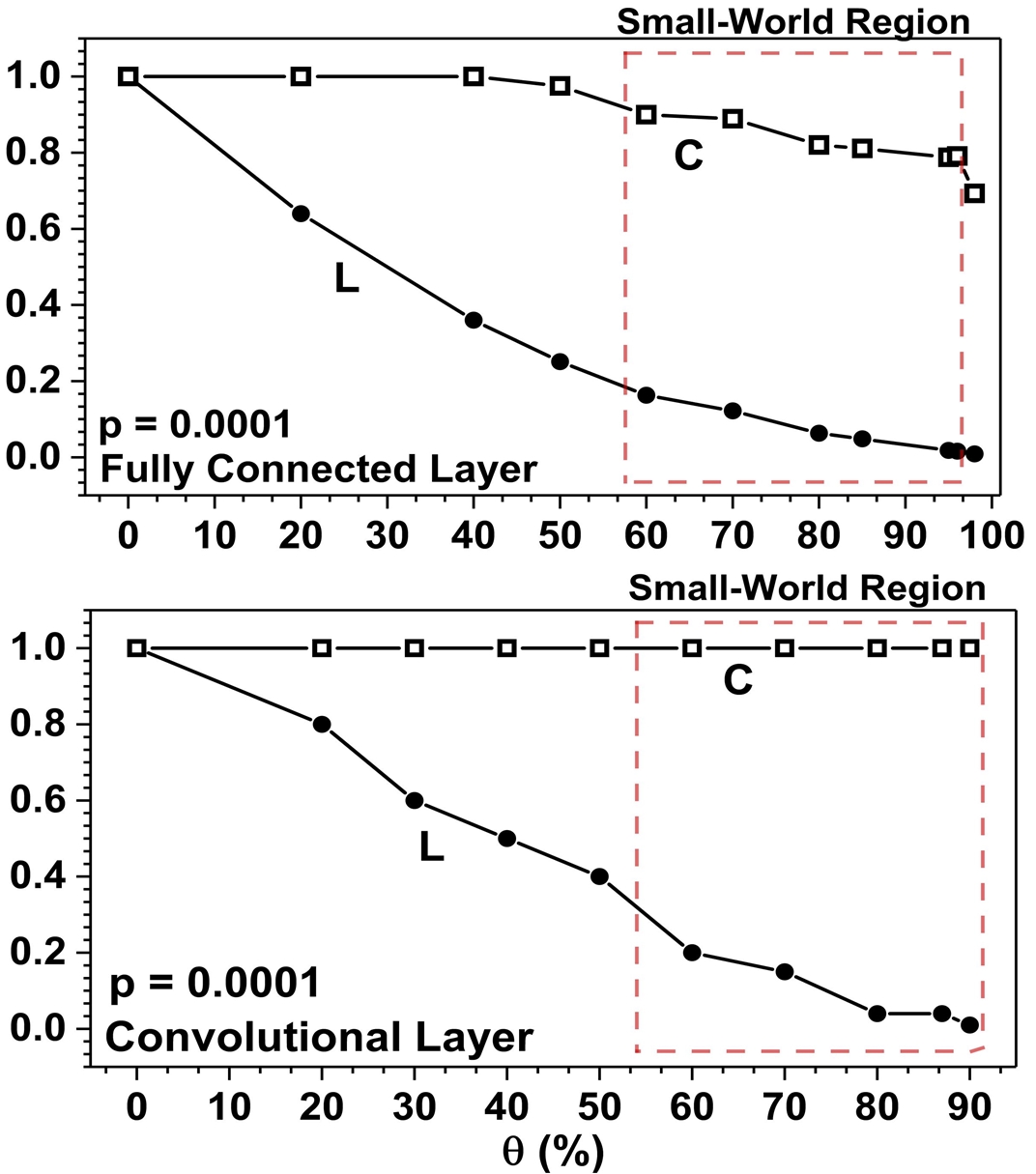}
\end{center}
\vspace{-0.3cm}
\caption{Variation in L and C for the fully connected and convolutional layer of a Small-World VGG-16 on CIFAR-10 with increase in sparsity $\theta$ of the layer. The curve shows high clustering coefficient and low Characteristic Path Length at high $\theta$ values for $p=0.0001$.}
\label{LC_final}
\end{figure}
\subsection{Randomness and Sparsity}
\begin{figure}[!t]
\vspace{-0.3cm}
\begin{center}
\includegraphics[width=1\columnwidth]{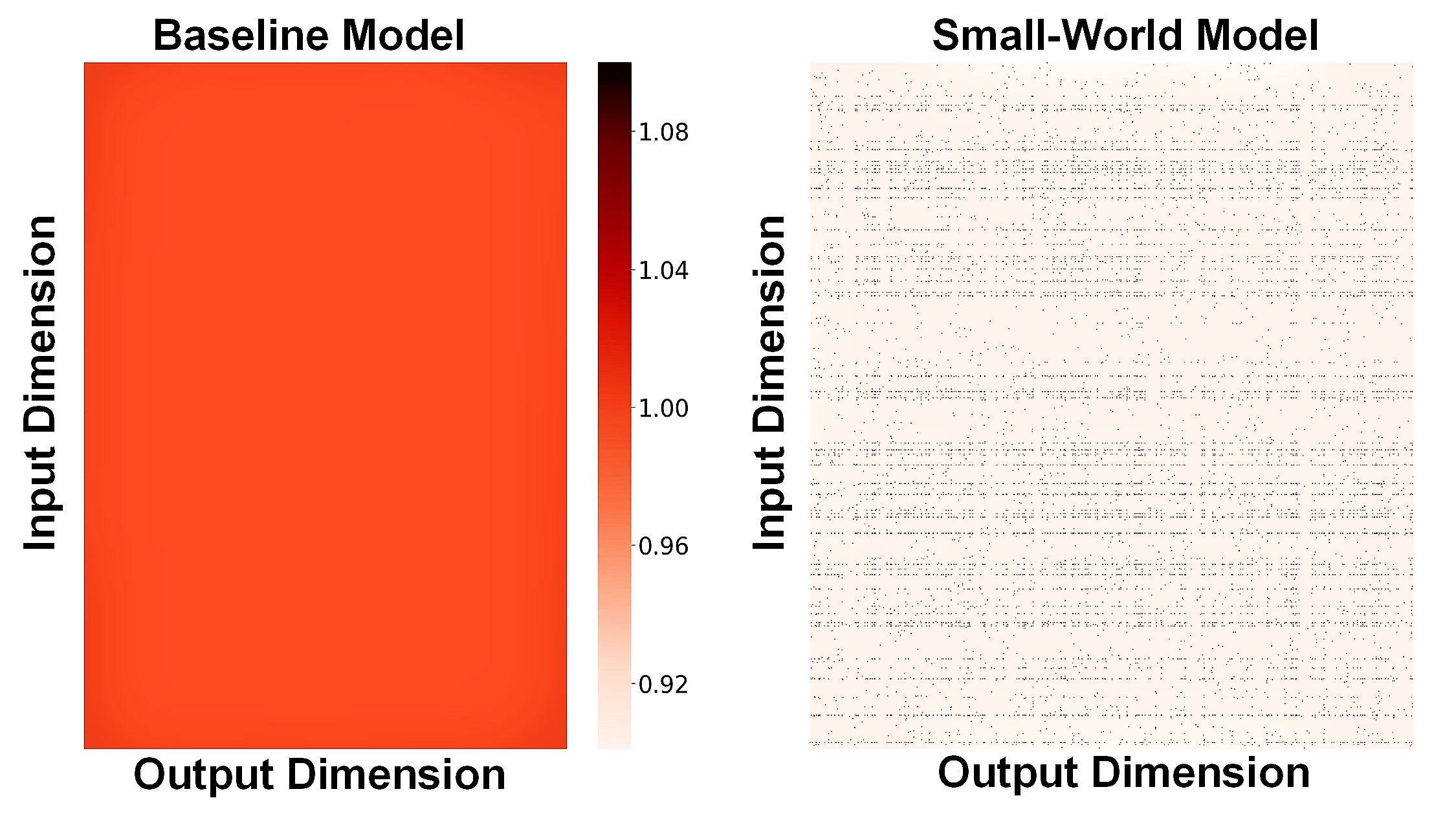}
\end{center}
\vspace{-0.2cm}
\caption{The heat map of the FC layer of VGG-16 on CIFAR-10 for both baseline (left) and Small-World pruned clustered model (right).}
\label{heat}
\end{figure}

\begin{figure}[h]
\begin{center}
\includegraphics[width=0.9\columnwidth]{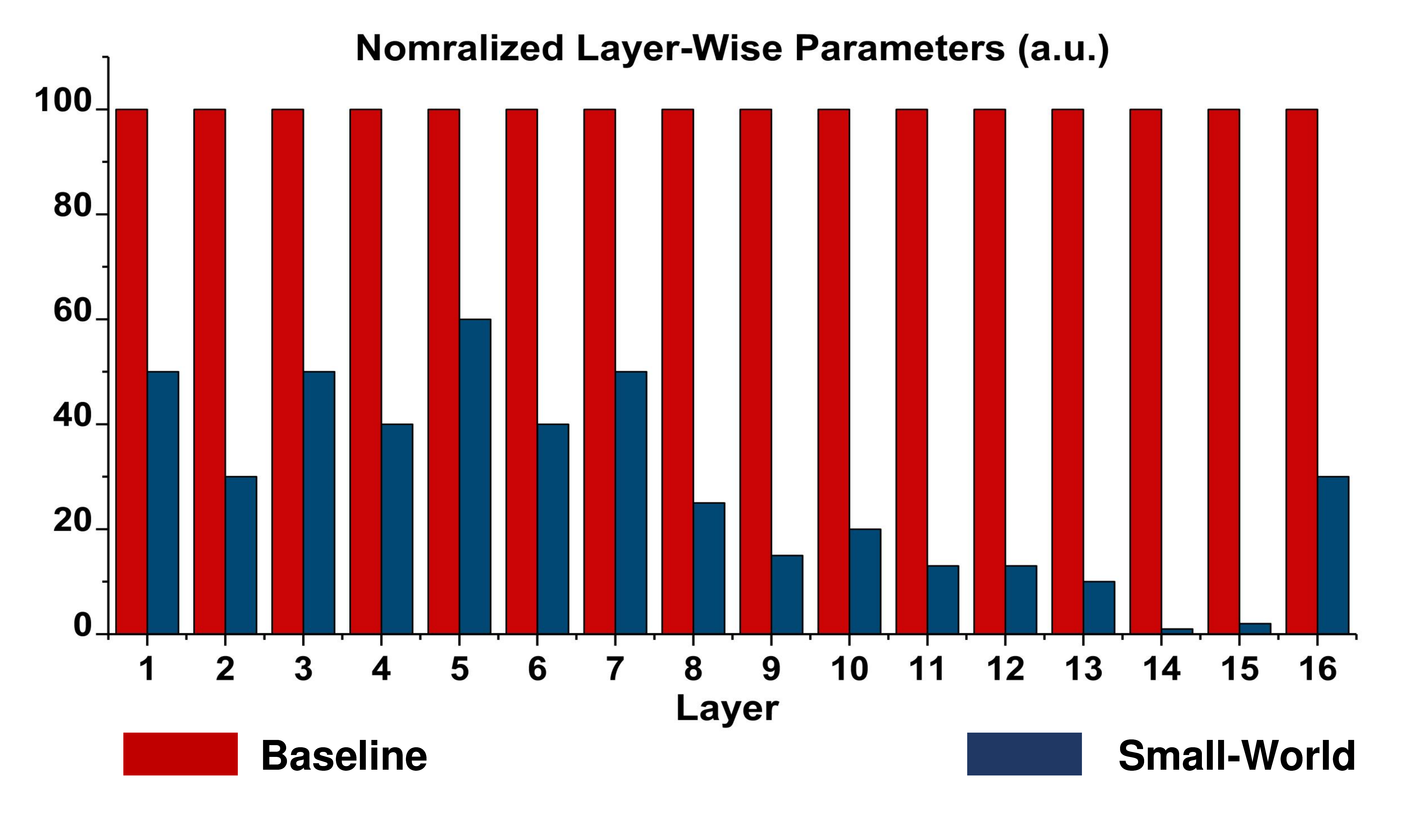}
\end{center}
\vspace{-0.3cm}
\caption{Layer-wise parameter distribution of the sparse Small-World based VGG-16 model compared to the baseline. Significant reduction in parameters is achieved, reducing both memory and interconnection cost.}
\label{interconnect}
\vspace{-0.25cm}
\end{figure}

Figure.~\ref{LC_final} shows the variation of L and C with increase in sparsity for a random probability value $p=0.0001$. The curve is shown for both FC and convolutional layers of the VGG-16 network. On training a Small-World modeled sparse network we achieve a model with high C and a low L for both type of layers. We argue that for higher layers the clustering is lower due to feature independence while for lower layers all features are mixed and hence all the feature maps contribute to the classes. Figure.~\ref{LC_final} denotes the Small-world region in the graph where C is high and L is low showing that our approach produces a well clustered model with high local density and global sparsity as shown in Figure.~\ref{heat}.
The heat map for the FC layer of Small-World modeled VGG-16 network shows local density for the parameters with global sparsity across the locally dense areas. This provides a platform for hardware efficiency with a cross-bar structure depicting the locally dense connections and long range connections cross them depicting the globally sparse connections.
We argue that with a sparse Small-World structure having high clustering and low connection density, the number of parameters can be reduced dramatically. This helps in the reduction of the number of interconnections on hardware. Figure.~\ref{interconnect} shows the normalized distribution of the interconnect density for VGG-16 network. We find that for higher layers the density of interconnections is <20\% of baseline model while for the lower layers it is of the order of 40-60\%. We argue that this is due to the similarity between Small-World property and interconnection distribution where for lower layers the features are more correlated resulting in higher clustering (C) and connection density (L) as compared to higher layers, resulting in higher hardware cost.
\section{Results}
In this section we present the results that summarize the advantages of the Small-World based pruning approach.
\subsection{Random Probability Estimation}
\begin{figure}[!b]
\begin{center}
\includegraphics[width=1\columnwidth]{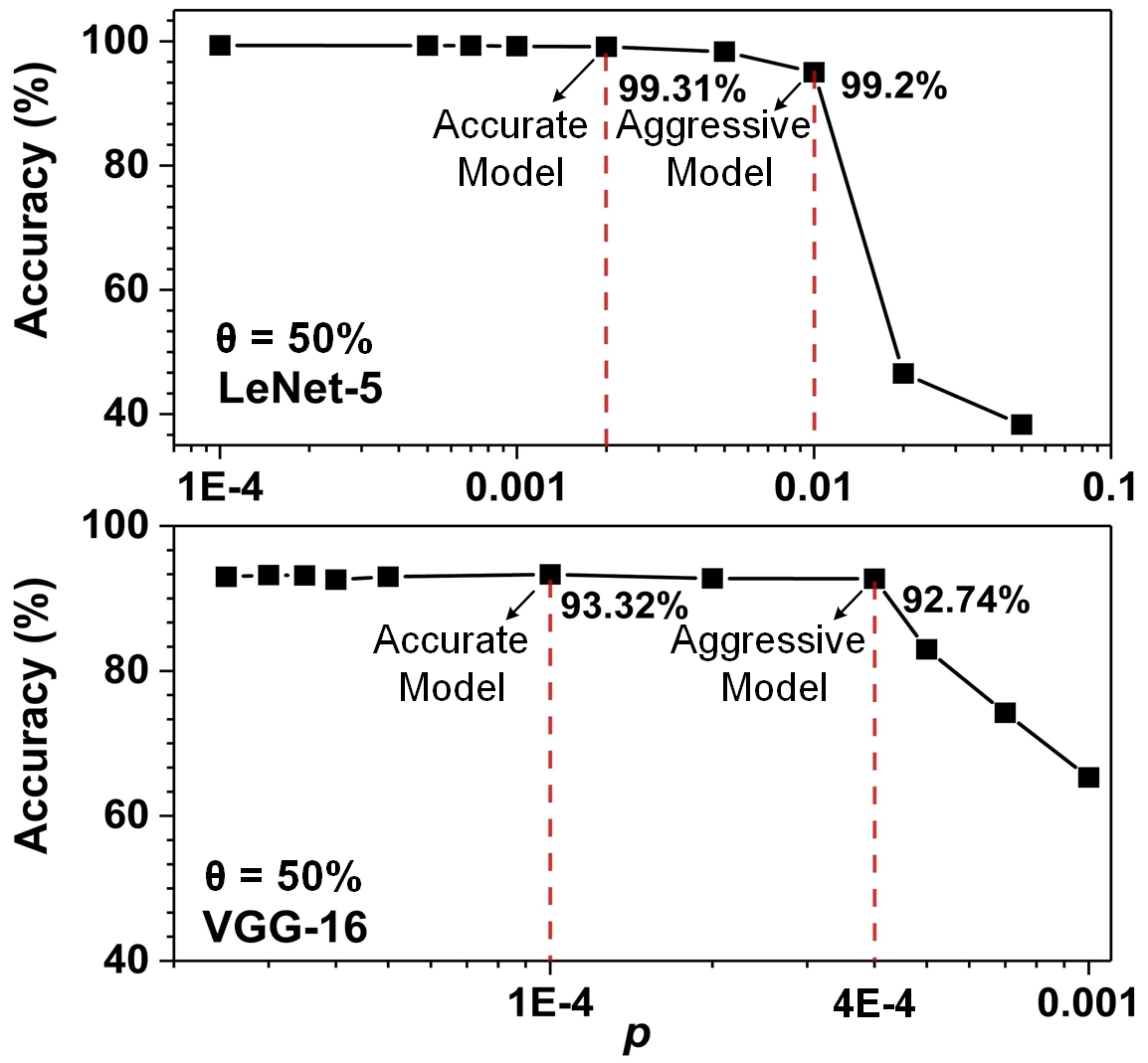}
\end{center}
\vspace{-0.4cm}
\caption{Estimation of maximum randomness that can be introduced into the network by extracting corresponding accuracy; for LeNet-5 on MNIST \textit{p=0.001} (top), for VGG-16 on CIFAR-10 \textit{p=0.0001} (bottom).}
\label{rand}
\end{figure}

Small-World Neural network model is defined by the randomness injected into the network based on the network structure properties. Figure.~\ref{rand} gives the variation of accuracy with increase in randomness for the network at $\theta$=50\%. We choose an optimum $p$ based on the accuracy metric for the network. For LeNet-5 on MNIST we choose $p$=0.001 due to better accuracy as compared to $p$=0.01. The sudden drop in accuracy at $p$=0.01 depicts the increase in randomness causing the network clustering property to fail similar to Figure.~\ref{lcthresh}. A similar approach is adopted for VGG-16 on CIFAR-10. 
\begin{table}[]
\begin{center}
\resizebox{\columnwidth}{!}{ 
\begin{tabular}[!t]{|l|l|l|l|ll}
\hline
\textbf{Model} & \textbf{Accuracy \%} & \textbf{Param.} & \textbf{Pruned \%} \\ \hline
LeNet-5& & &\\ \hline
Baseline & 99.29\% & 431K & - \\
Pruning \cite{hu2016network} & 99.26\% & 112K & 74.00\%\\
Pruning \cite{hansong} & 99.23\% & 34.48K & 92.00\%\\
Small-World-Accurate & \textbf{99.18\%} & \textbf{9.9K} & \textbf{97.70\%}\\
Small-World-Aggressive & \textbf{98.4\%} & \textbf{8.48K} & \textbf{98.04\%}\\\hline
VGG-16 Version-1 \cite{vgg-15}& & &\\ \hline
Baseline  & 92.93\% & 15.3M & - \\
Pruning \cite{HaoLi} & 93.4\% & 5.4M &  64.00\%\\
Small-World & \textbf{93.1\%} & \textbf{5.3M} &  \textbf{66.00\%}\\\hline
VGG-16 Version-2& & &\\ \hline
Baseline & 93.35\% & 41.5M & - \\
Small-World-Accurate & \textbf{93.24\%} & \textbf{3.74M} & \textbf{90.80\%} \\
Small-World-Aggressive & \textbf{92.72\%} & \textbf{3.4M} & \textbf{91.58\%} \\\hline
\end{tabular}
}
\vspace{0.2cm}
\caption{Overall result for LeNet-5 on MNIST and VGG-16 on CIFAR-10. Our algorithm provides a more structured approach and an effective reduction in model size as compared to other pruning works.}
\vspace{-0.5cm}
\label{overallresult}
\end{center}
\end{table}
\subsection{Accuracy and Model Compactness}
\begin{figure}[!b]
\begin{center}
\includegraphics[width=1\columnwidth]{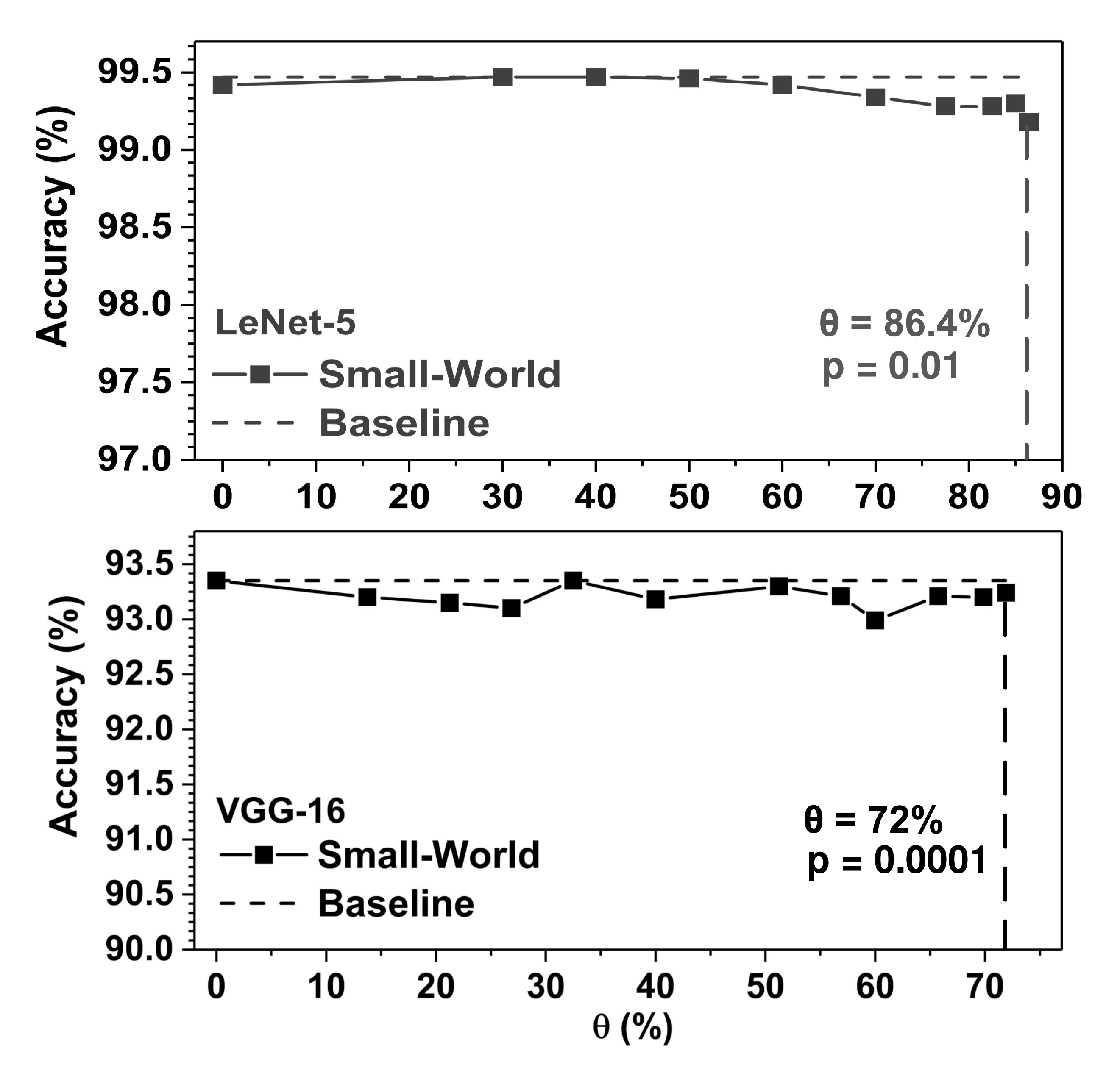}
\end{center}
\vspace{-0.5cm}
\caption{Accuracy variation for various sparsity configurations of the Small-World model of LeNet-5 on MNIST and VGG-16 on CIFAR-10.}
\label{acc_final}
\vspace{-0.3cm}
\end{figure}

The final model accuracy for LeNet-5 on MNIST and VGG-16 on CIFAR-10 are shown in Figure.~\ref{acc_final}. The accuracy is comparable to the baseline accuracy with a significant reduction in parameters. This provides a compact and accurate model through the Small-World pruning approach.
Table.~\ref{overallresult} summarizes the overall result of the Small-World based pruning approach. Our approach achieves a model compression of 97.7\% for LeNet-5 on MNIST, higher than existing methods. The number of parameters achieved for the Small-world model of LeNet-5 is lesser than that of \cite{hansong} for a similar accuracy. For Modified VGG-16 \cite{vgg-15} on CIFAR-10 we obtain lower number of parameters for a near baseline accuracy as compared to \cite{HaoLi}. For regular VGG-16 on CIFAR-10 we achieve a 90.8\% reduction in number of parameters for 93.24\% accuracy.
A locally dense, globally sparse structure provides a promising platform for efficient hardware implementation of Deep Neural Networks. In this work, we propose a novel method based on the Small-World networks to produce such a compact, clustered and accurate model.\\ 
The efficiency of this method is derived from the randomness induced into the network before training, which in turn improves the sparsity and clustering. The pre-defined sparse structure aids the deign of efficient hardware by reducing memory and interconnect cost. This approach produces clustered sparse models which provide structured sparsity to the final network which can model efficient hardware architectures in image recognition applications. We achieve a clustered locally dense globally sparse accurate model with 97.7\% reduction in parameters for LeNet-5 on MNIST and 90.8\% for VGG-16 on CIFAR-10 datasets.
\section{Acknowledgement}
This work was supported by C-BRIC, one of six centers in JUMP, a Semiconductor Research Corporation (SRC) program sponsored by DARPA.
\bibliographystyle{ieeetr}
\bibliography{references}
\end{document}